# High performing ensemble of convolutional neural networks for insect pest image detection

Loris Nanni[1*], Alessandro Manfè[1], Gianluca Maguolo[1], Alessandra Lumini[2] and Sheryl Brahnam[3]

[1]University of Padova, via Gradenigo 6, Padova 35131, Italy
2DISI, University of Bologna, via dell'Università 50, Cesena, Italy,
[3] Missouri State University, 901 S National Ave, Springfield, MO 65897, USA

**Abstract**

Pest infestation is a major cause of crop damage and lost revenues worldwide. Automatic identification of invasive insects would greatly speedup the identification of pests and expedite their removal. In this paper, we generate ensembles of CNNs based on different topologies (ResNet50, GoogleNet, ShuffleNet, MobileNetv2, and DenseNet201) altered by random selection from a simple set of data augmentation methods or optimized with different Adam variants for pest identification. Two new Adam algorithms for deep network optimization based on DGrad are proposed that introduce a scaling factor in the learning rate. Sets of the five CNNs that vary in either data augmentation or the type of Adam optimization were trained on both the Deng (SMALL) and the large IP102 pest data sets. Ensembles were compared and evaluated using three performance indicators. The best performing ensemble, which combined the CNNs using the different augmentation methods and the two new Adam variants proposed here, achieved state of the art on both insect data sets: 95.52% on Deng and 73.46% on IP102, a score on Deng that competed with human expert classifications. Additional tests were performed on data sets for medical imagery classification that further validated the robustness and power of the proposed Adam optimization variants. All MATLAB source code is available at https://github.com/LorisNanni/.

# 1 Introduction

Worldwide pest infestation is a major cause of crop and machine damage and, as a consequence, considerable reductions in grower revenues and economic growth [1, 2]. Invasive pests in the US, for example, were reported in 2016 to be racking up losses totaling at least seventy billion USD per year [3]. Rather than sending experts into the field, a common low-cost method for monitoring and controlling pest populations has been to lure insects into traps with pheromones, color, and light. Once specimens are collected in this manner, they must be identified and counted, which is an expensive, time-consuming task requiring taxonomic expertise. Automatic identification of invasive insects from trap images is a solution that offers the prospect of increased accuracy, reduced costs, and large-scale execution. However, species identification and the discrimination of pests that are harmful versus those that are safe is complicated by the small size of many insects, close resemblance between species, discoloration caused by soap and alcohol solutions contained within the traps, the pheromone cap, the loss of legs in sticky traps, vegetative debris, and insect clumping [4-6]. Some of these conditions can be overcome by developing systems that automatically and continuously inspect pests in the field, but the difficult problem of identifying insects within complex backgrounds—as they appear on different crops or on the ground with occlusions and variations in illumination—remains [2, 7-10].

In a recent 2020 survey [4] of the literature on pest detection published between 2015 and 2019, it was determined that 63% of the studies were based on convolutional neural networks (CNNs) [11], a powerful deep neural network model designed specifically to recognize visual patterns from images with minimal preprocessing, and 29% on more traditional feature-based approaches. An earlier survey published in 2018 [12] that focused on deep learners in pest identification found that only 42% of papers employed a CNN approach. These two surveys demonstrate that CNNs have become the classifier of choice in pest classification, a finding that is not surprising given the power of CNNs, which we highlight in section 4.

In this paper, we propose a method for classifying insects that is based on CNN and some new Adam optimization variants. During CNN training, millions of parameters are updated via a loss function, which attempts to minimize the difference between current outputs and the actual values in the training set. Traditionally, gradient descent (GD), especially the stochastic gradient descent (SGD) [13] has been the preferred method for minimizing an objective function. Although popular, SGD tends towards the convergence of suboptimal local minima when minimizing highly non-convex error functions due to the fact that the optimization landscape of SGD is not convex. To overcome this shortcoming, many variations of SGD have been proposed [14] [15-17]. To some degree, every SGD variant utilizes the momentum to direct the gradient [14-16]. In [14] and [18], for instance, optimization is guided by its own inertia. In AdaGrad [15] and Adadelta [16], which extends AdaGrad, the learning rate of parameters is decreased more with large partial derivatives than with smaller partial derivatives. The mechanism behind AdaGrad is the accumulation of previously squared gradients, unlike Adadelta, which keeps track of an exponentially decaying average of past squared gradients. Another SDG variant, Adam [18], not only stores an exponentially decaying average of past squared gradients but also an exponentially decaying average of past gradients; this information is used to decrease the learning rate of the parameters whose gradient changes more frequently, most especially in cases where the gradient changes its sign. Adam is well known for its realization of low minima of the training loss and is now frequently used with CNNs [19].

Adam's excellent ability to find low minima, however, fails to translate into better performance compared with SGD [20]. As a consequence, many new variants of Adam have been introduced that aim at increasing its effectiveness [17, 21, 22]. In [21], for example, the authors propose Nadam, which inserts into Adam the Nesterov momentum. More recently, new

Adam variants have been proposed that are based on the difference between present and past gradients, where the step size is adjusted for each parameter. In [22], AMSGrad curtails the step size to prevent it from increasing, while in diffGrad [17], which has obtained state-of-the-art results, the step size for every parameter is made proportional to the change in the gradient. In [23], the performance of different Adam variants is compared with SGD and shown to introduce diversity in ensembles of CNNs. In addition, the authors propose two new Adam optimization methods: 1) DGrad, which is a variant of diffGrad that uses a moving average of the squares of parameter gradients, and 2) two adjustments of DGrad that apply different cyclic learning rates [24].

In this study, we propose two new variants of DGrad [23]: Exp, which introduces a scaling factor, and ExpLR, which varies Exp by adding an extra step when calculating the final learning rate. Sets of CNNs are then trained on two insect benchmarks (Deng [9] and IP102 [25]) using these and several other Adam variants. Fusions of these partially independent classifiers are compared and evaluated to discover those combinations that produce the best classification results. Our best method is shown to achieve state of the art on both data sets. Additional tests are performed on some medical data to further validate the usefulness of the proposed Adam optimization variants.

The organization of the remainder of this paper is as follows. In section 2, a short review of related papers in pest classification is provided, followed, in section 3, by a brief introduction to the CNN topologies investigated in this study. In section 4, the Adam optimization variants tested in this work, along with the new ones proposed here, are detailed. In section 5, the benchmark data sets are described, and experimental results are presented. The paper concludes with some final remarks and some suggestions for future research.

## 2 Related Work on Insect classification

Several pest data sets for supervised learning are available. Early collections of pest images were small by today's standards, ranging in size from 200 samples divided into ten classes [26] to 1440 samples distributed into twenty-four balanced classes [8]. The popular Deng benchmark [9], sometimes referred to as SMALL, contains 563 samples in ten classes. These early data sets were geared towards training traditional classifiers on handcrafted features extracted from images. In [26], for example, the authors extracted Scale Invariant Feature Transform (SIFT) [27], Speeded-Up Robust Features (SURF) [28], and Histograms of Oriented Gradient (HOG) [29] descriptors that were trained on a Support Vector Machine (SVM) [30]. Likewise, in [9], the authors trained SVM on SIFT features and discovered regions of interest (ROI) using saliency maps [31], which detect ROI by identifying regions that are notably different from neighboring ones. Rather than using handcrafted features, the authors in [8] employed a multiple-task representation combined with multiple-kernel learning.

With rapidly increasing advancements in data capturing methods and storage systems, data size expectations increased exponentially (see [32] for a discussion on the surge in data size expectations in medical imaging). Deep learning methods are also fuelling the desire for larger data; CNNs, for instance, require massive data sets to reduce overfitting and increase performance. As noted in the introduction, most contemporary studies on pest classification are based on CNNs [4]. One of the first to use a CNN for pest identification was [33], where the researchers contended with the small size of the Deng data set using transfer learning applied to a version of AlexNet [11]. One of the contributions offered in [33] was a comparison of their CNN system with the accuracy rate of six human experts (AlexNet in that study outperformed four of the six experts on the Deng data). In [34], the authors calculated a saliency map and trained a version of AlexNet on a data set of 5000 images they collected on Google, Naver, and FreshEye. In [35], the authors build a data set of 30000 images of pests divided into eighty-two classes that were used to train AlexNet. Investigations into classifying pests with deep learning methods took a leap forward with the introduction of the IP102 benchmark [25]. This data set was carefully designed and contains over 75000 pest images (19000 annotated with bounding boxes) categorized into 102 classes. Both the data set and several pretrained CNNs (AlexNet, GoogleNet [36], VGGNet [37], and ResNet [38]) on IP102 are available to researchers.

Within the last couple of years, the three most popular CNN models for insect classification have been versions of VGGNet [5, 39-46], ResNet [5, 41, 44, 47-50], and MobileNet [44, 47, 51, 52], and most studies now compare performance using several CNNs evaluated on more than one data set. In [5], the authors proposed a model for automatic feature extraction and learning that was evaluated against a set of pretrained (on ImageNet) versions of AlexNet, ResNet, GoogleNet and VGGNet fine-tuned on the National Bureau of Agricultural Insect Resources (NBAIR) dataset of 40 classes of field crop insect images and the Xie1 (24 classes) and Xie2 (40 classes) data sets [8]. To prevent overfitting, simple data augmentation methods (reflection, rotation, translation, and scaling) were applied. In [53], new reuse feature residual blocks were stacked to generate a reuse residual network applied to ResNet and evaluated on IP102 (55.24% accuracy) and a couple of other benchmarks intended for different classification tasks. In [10], the authors introduced a method for pest classification that fused a set of CNNs (AlexNet, GoogleNet, DenseNet, ShuffleNet, and MobileNetv2) trained on images preprocessed using three saliency image methods that resulted in nine additional images for each original image in the data sets. The best ensemble reported in [10] obtained excellent performance on both the Deng (92.43% accuracy) and IP102 (61.93% accuracy) data sets, the two data sets used for comparison in this study. Finally, in [44], sets of pretrained (on ImageNet) CNNs (VGG-16, VGG-19, ResNet-50, Inception-V3, Xception, MobileNet, SqueezeNet) were modified and fine-tuned on Deng, IP102, and the D0 pest data set [42] of 4508 images divided into forty classes. An ensemble (SMPEnsemble) of the best three CNNs (Inception-V3, Xception, and MobileNet) in [44] were fused with a sum of maximum probabilities strategy and weighted voting. A second ensemble (GAEnsemble) had weights determined by a genetic algorithm. GAEnsemble obtained state-of-the-art on the Deng (95.16% accuracy) and IP102 (67.13% accuracy) data sets.

## 3 Convolutional Neural Networks (CNNs)

Although somewhat disregarded when first proposed due to their excessive computational costs, CNNs [11] have become, in the age of relatively cheap GPUs, one of the most popular deep learning models. Unlike classic neural networks, CNNs are composed of three specialized layers: a convolutional layer, a pooling layer, and a fully connected layer. The convolutional and pooling layers function together as a feature extractor, with the convolutional layer extracting region features from the inputs using convolution operators and receptive fields; these features are then passed on to the pooling layer, which uses such operators as max pooling, average pooling, and min pooling to reduce the dimensionality of the features. One or more fully connected layers along with a softmax function form the classifier component of the CN that maps inputs to class labels. Early work on CNNs mainly focused on building deeper architectures, a stream of research that was primarily motivated by research in [37], which demonstrated that building CNNs with sixteen or more layers results in better performance.

This section briefly describes the five CNN models (ResNet50, GoogleNet, ShuffleNet, MobileNetv2, and DenseNet201) used in this study. Two models, ShuffleNet and MobileNetv2 are included not only for their speed and efficiency but also for their ability to work well on mobile devices, useful in the field.

*3.1 ResNet50 (RN)*

ResNet50 [38] is a residual learning network (ResNet) composed of forty-eight convolution layers, one max pooling, and one average pooling layer. ResNet is able to train many layers due to its groundbreaking skip connections, which is a feature transmission method that prevents gradient vanishing (see Section 4). ResNet50 has more than twenty-three million trainable parameters and is highly popular.

*3.2 GoogLeNet (GL)*

GoogleNet [36] is relatively small, composed of only twenty-two layers (twenty-seven if counting the pooling layers) and four million trainable parameters. GL is known for its inception layers, which allow the network to choose between many convolutional filter sizes in each block. These filters of different sizes can handle different patterns in images.

*3.3 ShuffleNet (SN)*

ShuffleNet [54], proposed to work on mobile devices, dramatically reduces the number of parameters and computation time. SN uses group convolutions and channel shuffle to make a fast and highly efficient network. The grouped convolutions are $1 \times 1$ and are designed to handle only a subset of channels in the hidden layer, which significantly reduces the number of multiplications required. The channels of the output are shuffled to facilitate communication between neurons.

*3.4 MobileNetv2 (MN)*

MobileNetv2 [55] is another lightweight convolutional network made for mobile applications. MN is composed of depth-wise separable convolutions that are often found in CNNs designed for mobile devices. Conceptually, these separable convolutions are convolutional layers where the 3D weight tensor is split into a 2D and 1D tensor, a division that reduces memory requirements. MobileNetv2 is noted for its utilization of inverted skip connections, which reduces the number of operations required by the network.

*3.5 DenseNet201 (DN)*

DenseNet201 [56] is a large, fully connected CNN with 201 layers and 20 million trainable parameters. Unlike ResNet, which uses an additive method to merge the previous layer to the next layer, DN concatenates the output. This network increases gradient flow and feature reuse and ranks among the best performing convolutional networks.

## 4 Adam Optimization

As indicated in section 4, one vein of CNN research in image recognition has focused on the exploration of new topologies generated by stacking different specialized layers, as in the case of ResNet50, Googlenet, Shufflenet, Mobilenet, and DenseNet. However, one drawback in neural networks with deep architectures is the vanishing gradient problem, so named because the gradient gradually vanishes as it passes from output to input, resulting in the weights of networks closer to the inputs being insufficiently trained. Another drawback to deep neural networks results from their large number of parameters which complicates the optimization space, making the search for optimal weights challenging to find. Thus, equally crucial to enhancing CNN architecture to maximize performance is discovering robust and stable optimization algorithms.

Traditional first-order optimization methods, such as GD, SGD, and Adam, are standard methods for training CNNs. Adam is more popular now than the other two methods because of its utilization of momentum to improve the network's ability to search for a solution [19]. The advantages offered by Adam, as mentioned in the introduction, have led to the development of many variants. This section provides mathematical details of Adam and the variants used in this study, including the two novel ones, Exp and ExpLR, introduced here for the first time.

*4.1. Adam*

Adam [18] calculates the adaptive learning rates for each parameter with an update rule that depends on the value of the gradient at time $t$ and the exponential moving averages, $m_t$ (the first moment) and $u_t$ (the second moment), of the gradient and its square.

The first and second moments are

$$m_t = \rho_1 m_{t-1} + (1 - \rho_1) g_t \tag{1}$$

and

$$u_t = \rho_2 u_{t-1} + (1 - \rho_2) g_t^2, \tag{2}$$

where $g_t$ is the gradient at time t. The square on $g_t$ is the component-wise square, and $\rho_1$ and $\rho_2$ are the hyperparameters of the exponential decay rate for the first moment (typically set to 0.9) and the second moment estimates (0.999).

Because both moments are initialized to 0, the moving averages calculated in the first few steps are small. To offset any bias, the following bias-corrected versions of the moving averages is proposed in Adam:

$$\widehat{m}_t = \frac{m_t}{(1 - \rho_1^t)} \tag{3}$$

and

$$\widehat{u}_t = \frac{u_t}{(1 - \rho_2^t)}. \tag{4}$$

Immediately apparent is that $g_t$ can have both positive and negative components and $g_t^2$, being squared, can only have positive components. This difference in the gradient and its square means that the gradient changes sign often, potentially causing the value of $\widehat{m}_t$ to become much lower than $\sqrt{\widehat{u}_t}$, a situation that produces a very tiny step size. The final update for each $\theta_t$ parameter in the network is calculated as

$$\theta_t = \theta_{t-1} - \lambda \frac{\widehat{m}_t}{\sqrt{\widehat{u}_t} + \epsilon}, \tag{5}$$

where $\lambda$ is the learning rate (typically 0.001), $\epsilon$ is a small positive number ($10^{-8}$) introduced to prevent division by zero, and all the operations are component-wise.

*4.2. AMSGrad*

AMSGrad [22] is an Adam variant that relies on the maximum of previously squared gradients, $\bar{\bar{u}}_t$, rather than the exponential average $\hat{u}_t$ in equation (4) to update parameters:

$$\bar{\bar{u}}_t = \max(\bar{\bar{u}}_{t-1}, u_t). \tag{6}$$

The final update for each $\theta_t$ parameter of the network is the same as equation (5) except for the replacement of $\bar{\bar{u}}_t$ for $\hat{u}_t$:

$$\theta_t = \theta_{t-1} - \lambda \frac{\hat{m}_t}{\sqrt{\bar{\bar{u}}_t} + \epsilon} \tag{7}$$

*4.3. diffGrad*

The Adam variant diffGrad [17] relies on the difference of the gradient, $\Delta g_t$, to set the learning rate and lock parameters into the global minima, an idea that is based on the observation that a reduction in gradient changes is indicative of a global minima. The goal is to produce large step sizes when the gradient is changing faster and smaller step sizes when the gradient is changing at a slower rate. The update function requires the absolute difference of two consecutive steps of the gradient, defined as

$$\Delta g_t = |g_{t-1} - g_t|. \tag{8}$$

The final update for each $\theta_t$ parameter remains the same as in equation (5) and has the same definitions of $\hat{m}_t$ and $\hat{u}_t$, as in equations (3) and (4). The learning rate, however, is the Sigmoid of $\Delta g_t$:

$$\xi_t = Sig(\Delta g_t) \tag{9}$$

$$\theta_{t+1} = \theta_t - \lambda \cdot \xi_t \frac{\hat{m}_t}{\sqrt{\hat{u}_t} + \epsilon}, \tag{10}$$

where $Sig(\cdot)$ is

$$Sig(x) = \frac{1}{1 + e^{-x}} \tag{11}$$

*4.4 DGrad*

DGrad [23] is a variant of diffGrad that considers the difference between two consecutive steps in terms of the absolute value of the gradient, $g_t$, and the moving average of the element-wise squares of the parameter gradients, $avg_t$, as expressed in equation (2). Thus,

$$\Delta ag_t = |g_t - avg_t|, \tag{12}$$

which can be normalized as

$$\Delta \widehat{ag}_t = \left(\frac{\Delta ag_t}{\max(\Delta ag_t)}\right). \tag{13}$$

The original definition of $\xi_t$ in diffGrad (see equation 9) is redefined in DGrad as

$$\xi_t = Sig(4 \cdot \Delta \widehat{ag}_t), \tag{14}$$

with $\Delta \widehat{ag}_t$ multiplied by 4 to increase the range of the output.

The final update for each $\theta_t$ parameter is the same as it is for the original diffGrad (see equation 10).

*4.5 Cos*

Cos (labelled Cos#1 in [23]) is a variant of DGrad that introduces a cyclic learning rate [24] that has been shown to improve classification accuracy and generally with fewer iterations. The periodic function cos(x) defines the range of variation in the learning rate $lr_t$ as

$$lr_t = \left(2 - \left|\cos\left(\frac{\pi \cdot t}{steps}\right)\right| e^{-0.01 \cdot (mod(t, steps)+1)}\right), \tag{15}$$

where mod is the modulo function, and $steps$ (set to 30) is the period. When $lr_t = 0$, its value becomes $9 \cdot 10^{-4}$.

As illustrated in Figure 1, $\xi_t$ in Cos is multiplied by $lr_t$ in addition to 4 and $\Delta \widehat{ag}_t$ in equation (14):

$$\xi_t = Sig(4 \cdot lr_t \cdot \Delta \widehat{ag}_t). \tag{16}$$

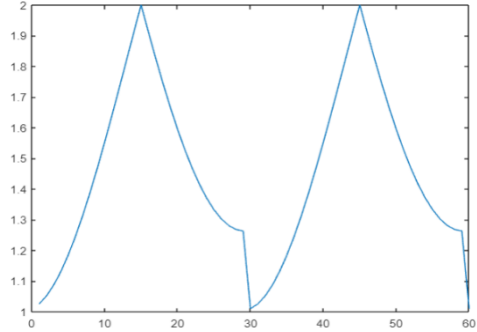

Figure 1. Cyclic learning rate: plot of $lr_t$ in the range of 1:2 × steps.

*4.6 Exp (new)*

Exp is a variant of DGrad that introduces a scaling factor $k$ in the learning rate that is applied to the absolute difference term $\Delta ag_t$. Higher values of $k$ (tested here with $k = 4$) make the curve steeper and more concentrated than do smaller values. The learning rate for Exp is

$$lr_t = k\Delta ag_t \circ e^{\circ(-k \cdot \Delta ag_t)}, \tag{15}$$

where ∘ is the Hadamard product defined as the element-wise matrix multiplication operator:

$$(A \circ B) = \begin{bmatrix} (a_{11} \cdot b_{11}) & \cdots & (a_{1j} \cdot b_{1j}) \\ \vdots & \ddots & \vdots \\ (a_{i1} \cdot b_{i1}) & \cdots & (a_{ij} \cdot b_{ij}) \end{bmatrix}, \qquad (16)$$

and where the Hadamard exponential is defined as the element-wise exponential:

$$e^{\circ A} = \begin{bmatrix} e^{a_{11}} & \cdots & e^{a_{1j}} \\ \vdots & \ddots & \vdots \\ e^{a_{i1}} & \cdots & e^{a_{ij}} \end{bmatrix}. \qquad (17)$$

As illustrated in Figure 2, the calculation of the final learning rate for the Exp DGrad variant is:

$$\xi_t = 1.5 \frac{lr_t}{max(lr_t)}. \qquad (18)$$

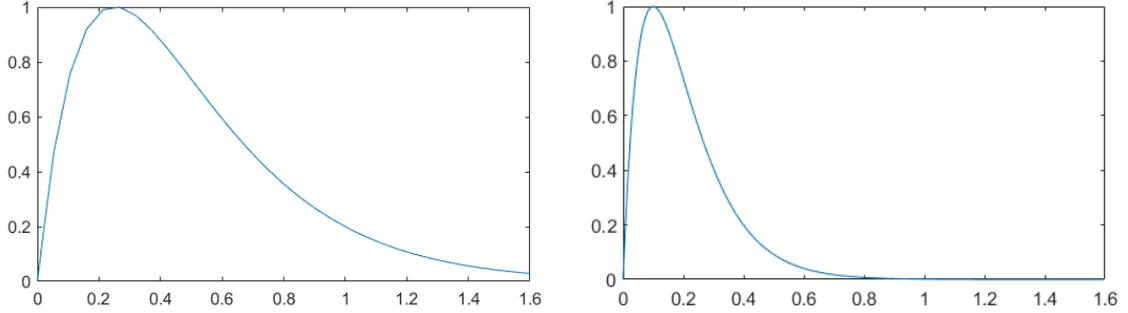

Figure 2. Exp using different values of k (k=4 left; k=10 right).

*4.7 ExpLR (new)*

ExpLR is a variation of Exp that adds a multiplicative term to the final learning rate. This term is given by the sigmoid function that maps the input parameter into [0.5,1]. In this case, our parameter is the normalized DGrad term $\Delta \widehat{ag}_t$ multiplied by a factor of 2 for better value distribution.

Given the definitions of $lr_t$ and $\widehat{lr}_t$ as

$$lr_t = \Delta ag_t \circ e^{\circ(-\Delta ag_t)} \qquad (19)$$

and

$$\widehat{lr}_t = 1.5 \frac{lr_t}{max(lr_t)}, \qquad (20)$$

and using the definition of the sigmoid function in equation (11), the final learning rate for ExpLR is

$$\xi_t = \widehat{lr}_t \circ Sig(2\Delta \widehat{ag}_t) = \frac{\widehat{lr}_t}{1 + e^{\circ(-2\Delta \widehat{ag}_t)}}. \qquad (21)$$

The sigmoid helps to normalize the output of the learning rate (see Figure 3) and further decreases values close to zero.

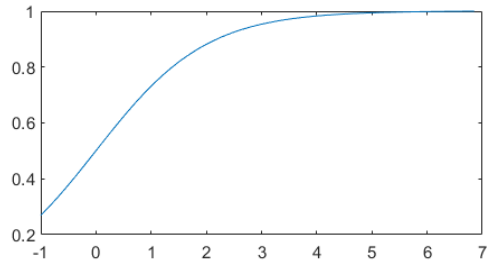

Figure 3. ExpLR: plot of $Sig(2\Delta \widehat{ag}_t)$

## 5 Experimental Results

*5.1 Data sets*

In the experiments below, we compare results on two pest data sets: Deng [9] and IP102 [25]. Both data sets, as illustrated in Figure 4, contain images of insects in the field.

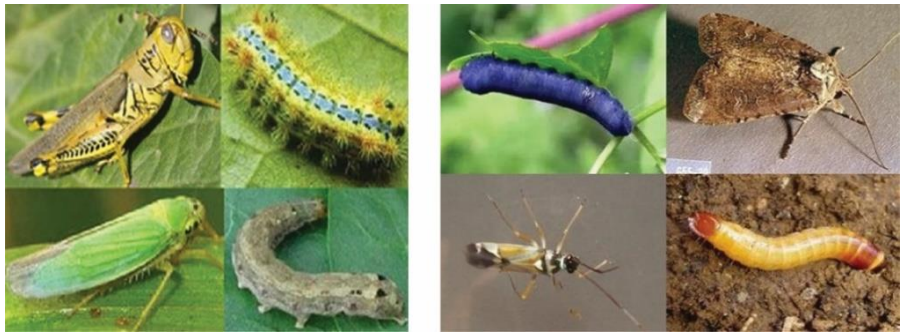

Figure 4. Image samples from Ding [9] (left) and from IP102 [25] (right).

The Ding data set contains images grouped into ten unbalanced classes of pests commonly found on plants between Europe and Central Asia and is a collection of 563 images taken from several online sources: Insert Images, IPM images, Dave's Garden, Mendeley Data, etc. The ten classes of pests and the number of samples in each class are the following:

1. Locusta migratoria (72): a type of locus in the family Arcrididae;
2. Parasa lepida (59): also known as the nettle caterpillar;
3. Gypsy moth larva (40): the immature larva of the Gypsy moth feeds on the leaves of more than 300 trees;
4. Empoasca flavescens (41): a genus of leafhoppers in the family Cicadellidae;

5. Spodoptera exigua (68): the small mottled willow moth or beet armyworm that attacks the leaves of cauliflower and other vegetables;
6. Spodoptera exigua larva (56): larvae of the small mottled willow moth;
7. Chrysocus chinensis (50): a genus of leaf beetles in the subfamily Eumopinae;
8. Atractomorpha sinensis (62): a pinkwinged grasshopper of the family Pyrgomorphidae that attacks many vegetables and rice;
9. Laspeyresia pomonella (65): also known as the Cydia pomonella mainly attacks fruits such as apples;
10. Laspeyresia pomonella larva (50): larva of the Cydia pomonella.

The split into training and test sets is the same as was reported in [9]. The training set consists of 20 images randomly extracted from each pest class for a total of 200 images, and the remaining 363 images form the test set. This split was performed five times with results averaged.

IP102 [25] is a large scale data set containing 75222 images divided into 102 classes of pests that are also categorized into a hierarchical taxonomy (developed in consultation with agricultural experts) that places each pest into a super-class reflecting the type of crop the pest preys on and a sub-class that labels the pests that ravage the crops. The classes contain images representing different stages in the insect's development, making the classification of each pest extremely difficult since the shape of a given insect can radically change as it matures, as in the case of a moth. As with the Ding data set, the IP102 collection was garnered online but taken not only from still images but also from video clips. The images were professionally annotated.

The classes of pests are highly unbalanced, ranging in sample size between 71 and 5740. In the experiments reported on this data set, the training set contained 45095 images and the test set 22169 images.

*5.2 CNN Settings*

When fine-tuning the CNNs, their last fully connected layer was replaced with a network whose output size was the number of classes of the given training data set. All the CNNs were trained using cross entropy as the loss function and with the following parameters:

- Batch size = 30 for the Deng data set, 50 for IP102 (40 if we have memory problem); we use a larger batch size in IP102 to reduce computation time;
- Number of epochs = 20;
- global learning rate = 0.001;
- gradient decay factor = 0.9;

- squared gradient decay factor = 0.999;
- loss function = cross entropy.

Because the Deng data set is small, simple data augmentation (reflection and random scale on both the axis) was performed. Data augmentation was not required for the much larger IP102 data set.

*5.3 Results*

In Table 1, we report classification performance on the Deng data set and in Table 2 on the IP102 data set using accuracy as the performance indicator. Rows represent the CNN architectures (DN, RN, GL, SN, and MN, as described in Section 3). Columns report results on each CNN using the following:

1. SA is the performance of the five single CNNs (DN, RN, GL, SN, and MN).
2. F_SA is the fusion by average rule of the SAs.
3. ENS_10 is the fusion by average rule of ten runs of a given CNN topology (DR, RN, GL, and MN).
4. F_ENS_10 is the fusion of the five ENS_10 results.
5. ENS_9+ENS-S is the fusion by average rule of nine runs (for fair comparison with ENS_10) of a given CNN topology (DR, RN, GL, and MN) combined with the saliency-based ensemble proposed in [10] and discussed in the Introduction. Thus, for each topology, 9+1 CNNs are fused.
6. F_ENS_9+ENS-S is the fusion of the five ENS_9+ENS-S results.
7. 2×ENS_9+ENS-S is the same as ENS_9+ENS-S except that ENS_9 has a weight of 2.
8. F_2×ENS_9 +ENS-S is the same as F_ENS_9+ENS-S except that F_ENS_9 has a weight of 2.
9. ENS_NEW is the fusion by average rule of the results of the ten runs on each CNN; a set of 10 trained using Exp and 10 using Exp_LR.
10. HERE is the weighted sum rule between F_SA (weight 1) and the fusion by average rule of the different ENS_NEW (i.e., the different new Adam topologies) with weight 2.

Except for classifiers using the Adam methods detailed in section 4, all the other approaches in Tables 1 and 2 are trained using SGD. When generating the sets of CNNs using SGD, a different data augmentation approach (random reflection, random rotation, and random translation) is applied randomly in order to perturb the training data for each epoch. In other words, each epoch uses a slightly different data set. The actual number of training images at each epoch does not change. The Adam-based approaches listed in Table 1 are the fusion by average rule of 10 runs for each CNN topology. Because of computational issues, for DN some Adam based approaches were not tested.

| Deng | DN | RN | GL | SN | MN |
|---|---|---|---|---|---|
| SA | 91.88 | 90.88 | 92.54 | 91.38 | 90.11 |
| F_SA | 94.70 | | | | |
| ENS_10 | 93.54 | 92.54 | 93.54 | 92.98 | 90.66 |
| F_ENS_10 | 94.70 | | | | |
| ENS_9+ENS-S | 93.26 | 93.09 | 93.81 | 92.98 | 91.99 |
| F_ENS_9+ENS-S | 94.81 | | | | |
| 2×ENS_9+ENS-S | 93.65 | 93.37 | 93.87 | 92.93 | 91.44 |
| F_2×ENS_9 +ENS-S | 94.86 | | | | |
| Adam | --- | 83.65 | 73.54 | 92.93 | 92.60 |
| DGRad | --- | 89.39 | 90.72 | 93.59 | 93.09 |
| Cos | --- | 89.94 | 90.11 | 93.81 | 92.60 |
| DiffGrad | --- | 90.55 | 91.16 | 93.81 | 93.76 |
| Exp (New) | 95.30 | 94.36 | 93.31 | 93.81 | 94.53 |
| Exp_LR (New) | 95.25 | 94.70 | 94.25 | 93.98 | 94.09 |
| ENS_NEW | 95.30 | 94.64 | 93.81 | 93.92 | 94.25 |
| HERE | **95.52** | | | | |
| [9] | 85.50 | | | | |
| [10] | 92.43 | | | | |
| SMPEnsemble [44] | 92.74 | | | | |
| GAEnsemble [44] | 95.16 | | | | |

**Table 2.** Accuracy rates for different methods trained on the Deng data set [9].

The following conclusions can be drawn from table 2:

- The new Adam-based approaches, Exp and Exp_LR, outperform both the other Adam-based methods and SGD (ENS_NEW outperforms ENS_10).

- The addition of saliency maps for data augmentation slightly improves the performance of CNNs trained with the original image (2×ENS_9+ENS-S outperforms ENS_10);
- The method labeled HERE obtains state-of-the-art performance on the Deng data set.

Because of the popularity of ResNet101 in the literature, we ran additional tests on this data set to compare ResNet101's performance on Adam (81.73) with its performance on the Adam variants DiffGrad (88.73), Exp (94.09), Exp_LR (95.27), and ENS_NEW (95.21). As with RN, the results on ResNet101 show that the new Adam variants perform better than Adam.

The confusion matrix obtained by our best approach (HERE) is reported in Figure 5. Nearly all the errors are due to a couple of classes that were also difficult for human experts to judge. Examples of some images our system and the experts had difficulty classifying are shown in Figure 6.

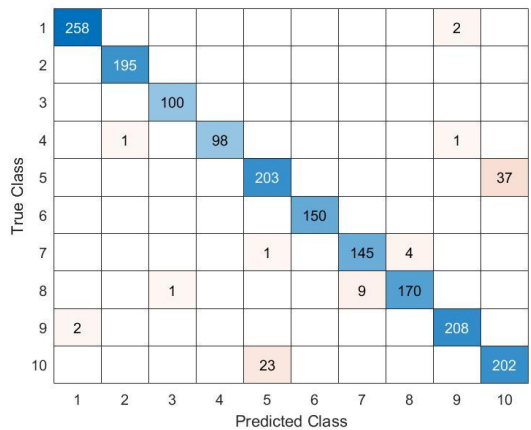

**Figure 4.** Confusion Matrix of HERE: the order of the classes is the same as in Table 1.

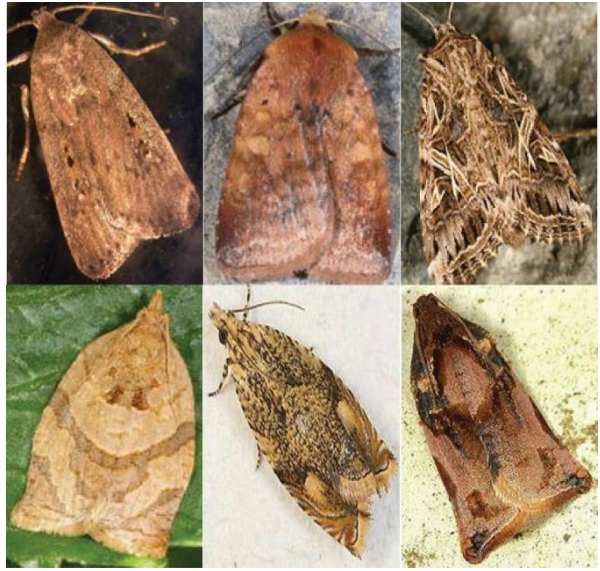

**Figure 5.** Difficult Images to Classify: first row are images of Spodoptera Exigua; last row of Laspeyresia Pomonella.

In Table 3, we report the performance obtained by six human experts in the Deng data set; it is interesting to note that our new approach comes close to matching the best experts.

| Expert      | 1    | 2    | 3    | 4    | 5    | 6    |
|-------------|------|------|------|------|------|------|
| Performance | 0.96 | 0.96 | 0.92 | 0.91 | 0.90 | 0.82 |

**Table 3.** Performance of Human Experts on the Deng Data set as reported in [33].

For some comparison with the literature discussed in section 2, in [9], the SVM trained on Deng obtained an accuracy of 85.5%, which falls far short of our best ensemble. In [33], the authors report an accuracy rate of 93.84% using transfer learning with AlexNet; however, a comparison with our approach is not fair since their system was tested on a partial and unbalanced version of Deng. Moreover, their protocol was a randomized training-test split where 70% of the images were used for a single training. Their technique retested with our protocol produces only 83.76% accuracy.

In Table 4, we report performance on the IP102 benchmark. To reduce computation time, only a subset of the approaches reported for the Deng data set are reported here. For the Adam methods, we do not perform 10 runs for each CNN method. Instead, we perform 7 runs for RN, GN and SN and 5 runs for MN (due to computational issue).

| IP102 | DN | RN | GL | SN | MN |
|---|---|---|---|---|---|
| SA | 69.74 | 65.40 | 64.11 | 57.36 | 63.21 |
| F_SA | | | 71.79 | | |
| Exp | --- | 69.80 | 67.79 | 68.80 | 71.12 |
| Exp_LR | --- | 70.74 | 68.13 | 67.76 | 71.02 |
| ENS_NEW | --- | 71.30 | 68.93 | 69.40 | 72.16 |
| Here | | | **73.62** | | |
| [25] | | | 49.50 | | |
| [53] | | | 55.24 | | |
| [10] | | | 61.93 | | |
| SMPEnsemble [44] | | | 66.21 | | |
| GAEnsemble [44] | | | 67.13 | | |

**Table 4.** Accuracy rates on the IP102 data set.

The results reported in Table 4 confirm the conclusions drawn from Table 2 that reported results on the Deng data set. Again, HERE outperforms the other approaches tested on the IP102 data set.

Because IP102 is highly unbalanced, two additional metrics (as reported in Table 5) were evaluated on this data set: F-score and G-means, with both computed as the weighted average of one versus all binary classifiers where the weight is equivalent to the fraction of samples in a given testing class. Of note is that HERE obtains state-of-the-art on both metrics.

| | F-score | G-mean |
|---|---|---|
| [25] | 0.401 | 0.315 |
| [53] | 0.541 | _____ |
| [10] | 0.592 | 0.755 |
| HERE | 0.729 | 0.847 |

**Table 5:** Comparison of the literature on IP102 using F-score and G-mean as the performance indicator.

Although our experiments demonstrate that our new Adam variants achieves the top performance compared with the literature on both the Deng and IP102 data sets, further evaluation on additional data sets in another domain is needed to confirm the strength and robustness of the new Adam variants. For this purpose, we tested our approach also on medical imaging data sets representing different classification tasks and labeled as follows:

- HeLa, which is the 2D HELA data set[*] [57], contains 832 grey-scale images of size 512×382 pixels. HeLa has 10 classes representing ten different organelles. The protocol in the literature used for testing is 5-fold cross-validation.

- BG, which is the Breast Grading Carcinoma data set[†] [58], contains 300 RGB images of size 1280×960 pixels divided into three classes representing grades 1-3 of invasive ductal carcinoma of the breast. The protocol in the literature used for testing is 5-fold cross-validation.

- LAR, which is the Laryngeal data set[‡] [59], contains 1320 patches images of size 100×100 pixels. This data set is divided into divided equally into four classes: IPCL (tissue with intrapapillary capillary loops), Le (tissue with leukoplakia), Hbv (tissue with hypertrophic vessels), and He (healthy tissue). The data set is divided into three subfolders intended for cross-validation.

In Table 6, the performance obtained by a stand-alone RN trained using different methods is reported. All the Adam-based methods are a fusion by average rule of seven CNNs (i.e., seven RN are trained on Exp then combined and seven on ExpLR then combined). The method named ENS_NEW is the fusion by average rule between Exp and ExpLR. SGD is the label given by the fusion of 14 RNs using SGD that were generated as above with the different augmentation methods. In this way, both SGD and ENS_NEW were built using the same number of CNNs. The method SGD+W× ENS_NEW is the weighted sum rule between SGD and ENS_NEW. In this test, the same data augmentations were applied to ADAM, as was the case for SGD.

---

[*] https://ome.grc.nia.nih.gov/iicbu2008/hela/index.html
[†] https://zenodo.org/record/834910#.YFsmIa9KiCo
[‡] https://zenodo.org/record/1003200#.YFsnR69KiCp

| accuracy | HeLa | BG | LAR | Deng |
|---|---|---|---|---|
| Adam | 74.30 | 89.67 | 96.29 | 83.37 |
| diffGrad | 94.88 | 91.67 | 95.91 | 90.55 |
| DGrad | 95.35 | 92.67 | 94.85 | 90.61 |
| Cos | 95.00 | 92.67 | 95.38 | 89.28 |
| Exp | 96.51 | 94.67 | 96.67 | 94.48 |
| ExpLR | 96.40 | 94.67 | 96.21 | 94.53 |
| ENS_NEW | **96.74** | **95.00** | **97.05** | **94.86** |
| SGD | 96.63 | 94.33 | 95.76 | 94.20 |
| SGD+ENS_NEW | 96.51 | **95.00** | 96.21 | 94.59 |
| SGD+2×ENS_NEW | 96.51 | **95.00** | 96.52 | 94.75 |

**Table 6**. Results on three other data sets that further validate the performance of the new Adam variants.

The following conclusions can be drawn from Table 6:

- The Adam variants, on average, perform better than the original Adam algorithm.
- Exp and ExpLR are the best-performing approaches across all data sets.
- The fusion by sum rule between Exp and ExpLR outperforms both the base approaches.
- Exp and ExpLR outperform SGD.

On LAR, Exp+ExpLR obtains an F-measure of 97.04. This result is better than both state-of-the-art performances reported in [59] and [60].

As final test, we ran experiments on the Kylberg Virus benchmark data set [61] (located at http://www.cb.uu.se/_Gustaf/virustexture/; accessed 08/27/2021). The virus benchmark has a total of 1500 Transmission Electron Microscopy (TEM) images (size 41×41) of viruses categorized into to fifteen types. The dataset is divided into two separate sets: 1) the object scale data set, where the radius of every virus is set to twenty pixels and 2) the fixed scale data set, where each virus image is represented so that each pixel corresponds to 1nm. Only the object scale data set is publicly available so widely reported in the literature. The fixed scale data set is proprietary and is unavailable for testing because of copyright issues.

To reduce computation time, we ran experiments on the object scale data set using a ten-fold protocol on Exp and ExpLR only (are best performing systems). The average accuracy of the single CNNs is: 84.97 for Exp and 85.21 for ExpLR; while the average accuracy of the fusion by average rule of the ten CNNs is 91.93 for Exp and 91.67 for ExpLR. The fusion by average rule of the ten CNNs trained with Exp and the ten CNNs trained with ExpLR obtains an accuracy of 92.53%. Clearly this ensemble strongly boosts performance.

Finally, in Table 7, we compare the performance of our best ensemble (Here) on the object scale virus benchmark with the best reported in the literature. As can be observed, our proposed method obtains state-of-the-art performance. In [61], the authors reported the performance on *fixed scale* data set. Because this data set is not available to other researchers, comparisons with [61] cannot be made.

| Here | [62] | [63] | [64] | [65] | [61] | [66] | [67] |
|------|------|------|------|------|------|------|------|
| 92.53 | 89.47 | 89.00 | 88.00 | 87.27 | 87.00* | 86.2 | 85.7 |

**Table 7.** Comparison with the literature Note: the method notated with * combines descriptors based on both *object scale* and *fixed scale* images.

**Conclusion**

In this paper, we combine CNNs based on different topologies (ResNet50, GoogleNet, ShuffleNet, MobileNetv2, and DenseNet201) and various Adam optimization methods for the task of pest identification. Two new Adam algorithms for deep network optimization based on the Adam variant DGrad are proposed that introduce a scaling factor in the learning rate that is applied to the absolute difference term. Sets of CNNs that vary in either data augmentation or the type of Adam optimization were trained on two benchmark insect data sets. Fusions were compared and evaluated using three evaluation metrics. The best performing ensemble, composed of ensembles of the CNNs and the new variants proposed here, is shown to achieve top results, compared to the literature, on both pest data sets. Additional tests were performed on benchmarks for medical imagery classification that further validated the robustness and power of the proposed Adam optimization variants: again, the ensemble approach presented here obtained state-of-the-art results.

In the future, we plan on applying active learning to the task of downloading a large number of images from the web to select a small number of new images to be labelled that might sensibly improve the classification performance.

All MATLAB source code can be downloaded from https://github.com/LorisNanni/.

# Acknowledgement

The authors wish to acknowledge the support of NVIDIA Corporation for their donation of the GPU used in this research.

# Author Contributions

Loris Nanni: Conceptualization, draft preparation, software, methodology. Alessandro Manfè: Gianluca Maguolo: Alessandra Lumini: and Sheryl Brahnam: Writing- Original, Visualization, Investigation, Writing- Reviewing and Editing